\documentclass[12pt]{article}
\usepackage{amsmath}
\usepackage[pdftex]{graphicx}
\usepackage{enumerate}
\usepackage{natbib}
\usepackage{url} 
\usepackage{amssymb}
\usepackage{amsthm}
\usepackage{color}
\usepackage{rotating}
\usepackage{multirow}

\DeclareMathOperator*{\argmin}{arg\,min}

\renewcommand{\baselinestretch}{1.5}
\def\cA{{\cal A}}

\def\cE{{\cal E}}

\def\cG{{\cal G}}
\def\cK{{\cal K}}

\def\smt{{\mbox{\tiny T}}}

\newcommand{\bX}{{\bf X}}
\newcommand{\bx}{{\bf x}}

\newcommand{\by}{{\bf y}}

\newcommand{\bu}{{\bf u}}

\newcommand{\bbeta}{\mbox{\boldmath{$\beta$}}}

\newcommand{\bfeta}{\mbox{\boldmath{$\eta$}}}
\newcommand{\bmu}{\mbox{\boldmath{$\mu$}}}

\newcommand{\bc}{\begin{center}}
\newcommand{\ec}{\end{center}}
\newcommand{\be}{\begin{equation}}
\newcommand{\ee}{\end{equation}}
\newcommand{\ba}{\begin{array}}
\newcommand{\ea}{\end{array}}
\newcommand{\bean}{\begin{eqnarray*}}
\newcommand{\eean}{\end{eqnarray*}}
\newcommand{\bea}{\begin{eqnarray}}
\newcommand{\eea}{\end{eqnarray}}
\newtheorem{Example}{\bf Example}

\newtheorem{theorem}{\bf Theorem}

\newcommand{\ben}{\begin{enumerate}}
\newcommand{\een}{\end{enumerate}}
\newcommand{\bed}{\begin{itemize}}
\newcommand{\eed}{\end{itemize}}

\newcommand{\bs}{\boldsymbol}
\newcommand{\blind}{0}

\begin{document}

\def\spacingset#1{\renewcommand{\baselinestretch}%
{#1}\small\normalsize} \spacingset{1}


\if0\blind
{
  \title{\bf Primal path algorithm for compositional data analysis}
  \author{Jong-June Jeon\\
    Department of Statistics \& Natural Science Research Institute 
    University of Seoul,\\
    Yongdai Kim \\
    Department of Statistics, Seoul National University,\\
    Sungho Won \\
    Department of Public Health Science, Seoul National University,    
    \\
    and Hosik Choi \\
    Department of Applied Statistics, Kyonggi University       
    }
  \maketitle
} \fi
\if1\blind
{
  \bigskip
  \bigskip
  \bigskip
  \begin{center}
    {\LARGE\bf Primal path algorithm for compositional data analysis}
\end{center}
  \medskip
} \fi

\bigskip
\begin{abstract}
Compositional data have two unique characteristics compared to typical multivariate data: the observed values are nonnegative and their summand is exactly one. To reflect these characteristics, a specific regularized regression model with linear constraints is commonly used. 
However, linear constraints incur additional computational time, which becomes severe in high-dimensional cases.
As such, we propose an efficient solution path algorithm for a $l_1$ regularized regression with compositional data. The algorithm is then extended to a classification model with compositional predictors. We also compare its computational speed with that of previously developed algorithms and apply the proposed algorithm to analyze human gut microbiome data.
\end{abstract}

\noindent%

\vfill

\newpage
\spacingset{1.45} 
	
\section{Introduction}\label{sec:intro}
In modern regression analysis, it is frequently observed that regression predictors consist of the proportions or relative ratios of certain values  rather than absolute values. For example, in analyzing air pollution data, the percentages of chemicals in the air are considered relevant predictors to identify the source of a pollutant \citep{lee2007heavy}. These types of  
proportional data, typically called compositional data, are widely used in geoscience \citep{buccianti2006compositional}, microbiology  \citep{Montassier2016}, and nutritional biochemistry \citep{leite2016applying}.
 By the definition of compositional data, all compositional predictors lie on the simplex and are thus linearly dependent.

\cite{aitchison1984log} proposed a regression model for compositional data as follows. 
Let ${\bf y} =(y_1, \cdots, y_n)^\top$ be a real valued response vector and $U = (\bu_1, \cdots, \bu_n)^\top$ the predictors, where $\bu_i$ is an element on the $p$-dimensional standard simplex. 
Instead of modeling $\bf y$ on $U$ directly, \cite{aitchison1984log} introduced a specific transformation of $U$, in which $\bx_i = \log(\bu_i)$ by the logarithm of each component of the vector, and proposed the following regression model with a constraint:
\bea
\label{eq:basic2}
{\bf y} &=& X \bbeta + \epsilon,\\
&&\mbox{subject to } {\bf 1}^\top \bbeta= 0, \nonumber
\eea
where $X = (\bx_1, \cdots, \bx_n)^\top$,
$\bbeta = (\beta_1, \cdots, \beta_p)^\top $, ${\bf 1} = (1, \cdots, 1)^\top \in \mathbb{R}^p$, and $\epsilon$ is a $n$-dimensional random vector with mean zero and finite variances. 
Subsequently, \cite{lin2014variable} adopted \eqref{eq:basic2} as a regression model with compositional data and proposed the $l_1$ regularized estimator of the regression coefficients, given by 
\bea \label{eq:comlasso}
\hat \bbeta(\lambda) &=&  \argmin_{\bs{\beta}}  \frac{1}{2}\|{\bf y} - X \bbeta\|^2_2 + \lambda \|\bbeta\|_1 \\
&& \mbox{subject to } {\bf 1}^\top \bbeta= 0. \nonumber 
\eea

However, the standard optimization algorithm for the $l_1$ regularized estimator is not directly applicable to solving (\ref{eq:comlasso}) due to constraint ${\bf{1}^\top} \bbeta = 0.$ 
For example, the coordinate-wise algorithm \citep{friedman2010regularization}, one of the most popular algorithms for $l_1$ regularization, cannot be used because 
the $l_1$ penalty function is non-separable \citep{tseng2009coordinate}. 
Alternatively, quadratic programming
\citep{brodie2009sparse,bondell2009simultaneous} or
the alternating direction method of multipliers (ADMM) algorithm can be used 
\citep{lin2014variable,fang2015cclasso}.
However, all these algorithms solve problem (\ref{eq:comlasso}) for a fixed $\lambda$ and do not provide a solution path. 

Various solution path algorithms for the $l_1$ regularized estimator with linear constraints have been developed as to provide a solution path by \citet{tibshirani2011solution}, \citet{Zhou2013}, and \citet{Zhou2014}.
 However, these algorithms have their own problems when applied
 to high-dimensional data. For example, the generalized lasso (genlasso) algorithm of \citet{tibshirani2011solution} does not generally provide 
 a solution path when $p>n$ and uses the additional $l_2$ penalty to solve this problem. However, adding the $l_2$ penalty not only increases computational complexity but also makes the corresponding solution path inaccurate (see Section 3 for details on this problem).
The algorithms of \citet{Zhou2013}, \citet{Zhou2014}, and \citet{gaines2018algorithms} solve general problems, including \eqref{ob2}.  However, a twice-differentiable loss function is required to solve the problem, which precludes further extensions such as a Huberized regression. 
Moreover, all aforementioned algorithms are not optimal for solving \eqref{eq:comlasso}, which has specific constraints.

The aim of this paper is thus to propose an algorithm to construct a solution 
path $\{\hat \bbeta(\lambda): \lambda \geq 0\}$, in which 
\bea \label{ob2}
 \hat \bbeta(\lambda) &=&\underset{  \bs{\beta} }{\mbox{argmin }}  L({\bf y}, X\bbeta)  + \lambda \|\bbeta\|_1\\
 &&\mbox{subject to }   {\bf d}_k^\top \bbeta_k = 0 \mbox{ for } k = 1, \cdots, K, \nonumber
\eea
Where $L:\mathbb{R}^n \times \mathbb{R}^n \mapsto \mathbb{R}$, $\bbeta = (\bbeta_{1}^\top, \cdots, \bbeta_{K}^\top)^\top \in \mathbb{R}^{p}$ with $\bbeta_k \in \mathbb{R}^{p_k}$, and ${\bf{d}}_k \in \mathbb{R}^{p_k}$. 
We let $L(\by, X\bbeta) = \sum_{i=1}^n l(y_i, \bx_i^\top \bbeta)$ and consider $l:\mathbb{R}\times \mathbb{R} \mapsto \mathbb{R}$ as 
the class of piecewise quadratic
loss functions \citep{rosset2007piecewise} including quadratic, Huberized, and quadratic hinge loss. In the regression model, $y_i\in \mathbb{R}$, and in the classification model, $y_i  \in \{-1,1\}$. Additionally, we introduce constant vectors ${\bf{d}}_k$ for $k = 1, \cdots, K$ to consider the more general $l_1$ constraint in the problem. For example, multiple compositional predictors can be included in the model.
We call the proposed algorithm the composite-lasso (comlasso).

The main contribution of this paper is clearly describing the event of violation for the Karush-Kuhn-Tucker (KKT) conditions corresponding to the linear constraint in \eqref{ob2}. We monitor the event and employ the monitoring result to determine the solution path. This enables us to seek the exact initial solution referred to in \citet{gaines2018algorithms} when $p>n$ and also provides an efficient way to find the exact activation set in the solution path.
To the best of our knowledge, this is the first paper to provide an exact description of the violation. Therefore, comlasso has the following advantages compared to other algorithms: it is computationally more efficient, especially for high-dimensional data, requires no modification by adding the $l_2$ penalty, and works with various loss functions other than square loss. In this paper, the intercept of the model is not considered for convenience. However, the proposed algorithm can be easily modified to include the intercept.

The remainder of the paper is organized as follows. Section 2 explains the KKT conditions
for problem (\ref{ob2}) and describes the solution path generation rule. Section 3 provides the results of the simulation study and the microbiome data analysis. Section 4 presents concluding remarks.

Finally, we define the notations used in this paper. For a positive integer $k,$ ${\bf 1}_k$ and ${\bf 0}_k$ denote the $k$-dimensional vector whose elements are $1$ and $0$, respectively. For an index set $\mathcal{A}\subset \{1, \cdots, p\}$, $v_{\mathcal{A}}$ denotes the subvector of $v$, whose elements are chosen by index set $\mathcal{A}$. Similarly $M_{\mathcal{A}}$ denotes the submatrix of $M$, whose columns are drawn according to index set $\cA$. When $\cA$ is singleton, we denote the index set by its element (e.g., $v_j$ and $M_j$ for $\cA  = \{j\}$).
Let $\cG_k$ for $k = 1, \cdots, K$ be the consecutive partition of $\{1, \cdots, p\}$, with $|\cG_k| = p_k$, where $|\cG|$ is the cardinality of set $\cG$. Let $k(j)\in \{1, \cdots, K\}$ be the membership of $j\in \{1, \cdots, p\}$, that is, $j \in \cG_{k(j)}$. We denote $L({\by}, X\bbeta)$ simply by $L(\bbeta)$. Finally, $\preceq$ is the elementwise inequality in the vector.

\section{Primal path algorithm for comlasso}

  The almost quadratic loss function $l:\mathbb{R}\times \mathbb{R} \mapsto \mathbb{R}$ is defined by $l(y,\bx^\top \bbeta)=a(r)r^2+b(r
 )r+c(r)$, where $a(r), b(r),$, and $c(r)$ are piecewise constants depending on $r$, which is a function of $y$ and $\bx^\top \bbeta$. In the regression model, $r$ is the residual, $y-\bx^\top \bbeta$, and in the classification model, $r$ is the margin, $y\bx^\top\bbeta.$ \cite{rosset2007piecewise} 
 proposed an algorithm for obtaining the piecewise linear solution path of the $l_1$ regularized optimization problem with the use of the above almost quadratic loss function. Essentially, piecewise linearity is derived from the condition that the second-order derivative of the loss function is locally constant.
 See examples \ref{EX:loss1} and \ref{loss:hinge} in the Appendix, where lists of popular almost quadratic functions are illustrated.
 Additionally, comlasso employs the property of the loss function and finds a solution set of \eqref{ob2} satisfying piecewise linearity. Therefore, comlasso is mainly based on conventional solution path algorithms.  
 However, it monitors the residuals or margins between $\by$ and $X\bbeta$ on $\bbeta \in \mbox{null}(D^\top)$ and exactly computes the dual parameters corresponding to the equality constraint $D^\top \bbeta = {\bf 0}_K$. This is the key step to implement the comlasso algorithm in a high-dimensional model.

 The comlasso algorithm consists of four steps: determining the first active coefficients with maximum $\lambda$, identifying the direction of solution path for the active coefficients, computing the step size of the solution path by monitoring the violation of KKT conditions, and updating the active coefficients and dual parameters. The last three steps are repeated until the algorithm cannot update the set of active coefficients anymore. 

\subsection{KKT conditions and initialization of comlasso}

The Lagrangian of primal problem \eqref{ob2} is given by 
\bea\label{problem:lag}
    L(\bbeta)+\lambda\|\bbeta\|_1 + \bmu^\top D^\top\bbeta,
\eea 
where $\bbeta = (\beta_1, \cdots, \beta_p)^\top$, $\bmu = (\mu_1, \cdots, \mu_K)^\top \in  \mathbb{R}^K,$ 
\bean
D = \left(\ba{cccc}
    {\bf d}_1 & {\bf 0}_{p_1} & \cdots & {\bf 0}_{p_1}\\
    {\bf 0}_{p_2} & {\bf d}_{2} & \cdots & {\bf 0}_{p_2}\\
    \vdots & \vdots & \vdots & \vdots\\
    {\bf 0}_{p_K} & {\bf 0}_{p_K} & \cdots & {\bf d}_{K}
    \ea \right),
\eean
and $\bmu$ is the dual parameter corresponding to the equality constraint. The KKT conditions are as follows: there exists a vector $(\bbeta, \bmu)$, so that 
\bea
\label{eq:KKT_active}
\nabla_{\cA} 
 L(\bbeta)+(D\bmu)_{\cA} +\lambda {\rm sign}(\bbeta_{\cA}) &=& {\bf 0}_{|\cA|}\\
\label{eq:KKT_inactive}
| \nabla_{\cA^c}  L(\bbeta) + (D\bmu)_{\cA^c} | &\preceq& \lambda\bf{1}_{|\cA^c|}\\
\label{eq:KKT_zerosum}
D^\top \bbeta &=& {\bf 0}_{K},
\eea
where $\cA=\{j: \beta_{j} \neq 0\}$, $\nabla  L(\bbeta)$ is the gradient vector of $L(\bbeta)$ and $\nabla_{\cA} L(\bbeta)$ denotes $(\nabla L(\bbeta))_{\cA}$. Eqs. \eqref{eq:KKT_active} and \eqref{eq:KKT_zerosum} are the stationarity and primal feasibility conditions, respectively. Eq. \eqref{eq:KKT_inactive} is a stationarity condition with the existence of dual parameter $\bmu$, which plays a key role in comlasso. Note that, if it exists, the dual parameter $\bmu$ is not necessarily unique.

Specifically, to describe the existence of $\bmu$, we introduce the dual feasible set of $\bmu$ for a given $\lambda$ and $\bbeta$: 
\bean
\cE_{\lambda}(\bbeta) = \{\bmu \in \mathbb{R}^K:|\nabla L(\bbeta) + D\bmu  |  \preceq \lambda {\bf 1}_p \}.
\eean
Particularly, let $\bbeta = {\bf 0}_p$; then, $\cE_{\lambda}({\bf 0}_p)$
is the feasible set of $\bmu$ corresponding to the initial solution of comlasso. If we let $\lambda_{\max} = \inf \{\lambda \in \mathbb{R}: \cE_{\lambda}({\bf 0}_p) \neq \phi\}$,  $\bbeta = {\bf 0}_p$ is a trivial optimal solution of \eqref{ob2} for all $\lambda \geq \lambda_{\max}$. However, for $\lambda <\lambda_{\max},$ $\bbeta = {\bf 0}_p$ is not the solution of \eqref{ob2} anymore, because the corresponding $\bmu$ does not exist.
Therefore, comlasso sets $\lambda = \lambda_{\max}$ as the initializing regularization parameter.

Obviously, $\cE_{\lambda}({\bf 0}_p)$ is not empty if and only if
\bean
\max_{j':j'\in \cG_k, d_{j'} \neq 0} \left( -\frac{\lambda}{|d_{j'}|}  - \frac{\nabla_{j'} L({\bf 0}_p)}{d_{j'}}  \right) 
\leq 
\min_{j:j\in \cG_k, d_j \neq 0} \left( \frac{\lambda}{|d_{j}|} - \frac{\nabla_{j} L({\bf 0}_p)}{d_{j}}   \right), 
\eean
for all $1\leq k \leq K$, where 
${\bf{d}} = ({\bf d}_1^\top, \cdots, {\bf d}_K^\top)^\top = (d_1, \cdots, d_p)^\top$. Note that ${\bf d}_k = (d_{j}: j \in \cG_k)$ and $\cG_k$ is as per the notation described above. Therefore, we have
\bea\label{lambda_max}
\lambda_{\max} = \max_{1\leq k \leq K}\max_{(j,j'): \substack{j,j'\in \cG_k,\\ d_j,d_{j'}\neq 0}}
\left(\frac{\nabla_{j'} L({\bf 0}_p)}{d_{j'}}  - \frac{\nabla_{j} L({\bf 0}_p)}{d_{j}}\right)\left(\frac{1}{|d_{j'}|} + \frac{1}{|d_{j}|}\right)^{-1}.
\eea
If there exists the unique triplet $(k,j,j')$ satisfying Eq. \eqref{lambda_max}, the $\mu _k$ satisfying \eqref{eq:KKT_zerosum} is uniquely given by $\lambda_{\max}/d_{j'} - \nabla_{j'} L({\bf 0}_p)/d_{j'}$ or $-\lambda_{\max}/{d_j} - \nabla_j L({\bf 0}_p)/{d_j}$.  Furthermore, the sign of $\beta_j$ and $\beta_{j'}$ is assigned by $\mbox{sign}(d_j)$ and $-\mbox{sign}(d_{j'})$, which will be explained in Section 3.3. The dual feasible set $\mathcal{E}_{\lambda}(\bbeta)$ will be used to re-check the violation of the KKT conditions. 

\medskip

\noindent \textbf{Remark}. \cite{gaines2018algorithms} proposed a method to find an initialized solution under the constrained lasso, which includes problem \eqref{ob2} as a special case. However, it is not clearly shown whether one or more coefficients can be simultaneously activated on the null space of $D^\top$. In problem \eqref{eq:comlasso}, two or more coefficients should be simultaneously activated in the initialization of the solution path algorithm. Additionally, all the signs of activated coefficients cannot be equal. 
When there are the multiple triplets in \eqref{lambda_max}, comlasso chooses only one of them and activates the corresponding coefficients.

\subsection{Finding the direction of the solution path}
\cite{rosset2007piecewise} proposed a toolbox for the linear solution path. We explain the piecewise linearity of the solution path based on this toolbox, considering the regression problem for convenience. To emphasize the dependence on $\lambda$, we denote the solution of \eqref{problem:lag} by $\bbeta(\lambda)=(\beta_1(\lambda), \cdots, \beta_p(\lambda))^\top$, and let the active variable and group variable index set be
$\cA = \{j:  \beta_j(\lambda) \neq 0\}$ and 
$\mathcal{K} = \{k: \|\bbeta_{\cG_k}(\lambda)\|_1 \neq 0\}$, respectively. By the KKT conditions in \eqref{eq:KKT_active} and \eqref{eq:KKT_zerosum}, we have the following two equations:
\bean
-\sum_{i=1}^n \left(2a(r_i(\lambda)) r_i(\lambda) + b(r_i(\lambda))\right) \bx_{i\cA}   + (D\bmu(\lambda))_{\cA} + \lambda \mbox{sign}(\bbeta_{\cA}(\lambda)) &=& {\bf 0}_{|\cA|}, \\
D^\top \bbeta(\lambda) &=& {\bf 0}_{K},
\eean
where $r_i(\lambda) = (y_i - \bx_{i \cA}^\top \bbeta_{\cA}(\lambda))$. 
We set a candidate of solutions $\bbeta(\lambda - \delta)$ with $\delta>0$ as 
\bean
\bbeta_{\cA}(\lambda - \delta) &=& \bbeta_{\cA}(\lambda) + \delta {\bf b}(\lambda),\\ 
\bbeta_{\cA^c}(\lambda - \delta) &=& {\bf 0}_{|\cA^c|}, \mbox{ and }\\
\bmu_{\cK}(\lambda - \delta) &=& \bmu_{\cK}(\lambda) + \delta {\bf m}(\lambda),
\eean
where ${\bf b}(\lambda)$ and ${\bf m}(\lambda)$ are the solutions of linear equation
\bea\label{eq:linear}
\left(\ba{c c}
H(\lambda) & D_{\cA}\\
(D^\top)_{\cA} & O
\ea\right)
\left(\ba{c}
{\bf b}(\lambda)\\
{\bf m}(\lambda)
\ea\right) =
\left(\ba{c}
\mbox{sign}(\bbeta_{\cA}(\lambda)) \\
{\bf 0}_{|\mathcal{K}|}
\ea
\right),
\eea
$H(\lambda) = \sum_{i=1}^n 2a(r_i(\lambda))\bx_{i \cA} \bx_{i \cA}^\top$, and 
$O$ is the $|\mathcal{K}| \times |\mathcal{K}|$ zero valued matrix. $({\bf b}(\lambda)^\top,{\bf m}(\lambda)^\top)^\top$ uniquely exists if 
$H(\lambda)$ is positive and definite and $D_{\cA}$ is full rank. It is easily verified that $\bbeta(\lambda - \delta)$ and $\bmu_{\cK}(\lambda - \delta)$ always satisfy
\bean
\nabla_{\cA} L(\bbeta(\lambda - \delta))+(D_{\cK}\bmu_{\cK}(\lambda - \delta))_{\cA} +(\lambda-\delta) {\rm sign}(\bbeta_{\cA}(\lambda - \delta)) &=& {\bf 0}_{|\cA|},
\eean
and $D^\top \bbeta(\lambda-\delta) = {\bf 0}_p$, if $\mbox{sign}(\bbeta(\lambda)) = \mbox{sign}(\bbeta(\lambda - \delta))$, $a(r_i(\lambda-\delta)) = a(r_i(\lambda))$, and $b(r_i(\lambda-\delta)) = b(r_i(\lambda))$ for all $i$. These two equations imply that $\bbeta(\lambda - \delta)$ and $\bmu(\lambda - \delta)$ are respectively the primal and dual solutions of 
\bea \label{eq:ob3}
&&\min_{\beta} L(\bbeta) + (\lambda - \delta)\|\bbeta\|_1\\
&&\mbox{subject to }D^\top \bbeta = 0, \nonumber
\eea
as far as there exists $\bmu_{\cK^c}(\lambda - \delta)$  so that
$$| \nabla_{\cA^c}  L(\bbeta(\lambda-\delta)) + (D\bmu(\lambda - \delta))_{\cA^c} | \preceq (\lambda-\delta)\bf{1}_{|\cA^c|}.$$
Particularly, the existence of $\bmu_{\cK^c}(\lambda - \delta)$ is verified by the dual feasible set $\mathcal{E}_{\lambda-\delta}(\bbeta(\lambda - \delta))$ defined in Section 2.1. Therefore, for $\bbeta(\lambda - \delta)$ and $\bmu(\lambda - \delta)$ with $\delta \geq 0$, we can categorize the violations of the KKT conditions and updating rules as follows:

\noindent \rule[0.5ex]{\linewidth}{1pt}
\bc
{\bf Rules for updating the activated primal and dual solutions}
\ec

\bed
\item[(C1)] termination condition: if $\lambda - \delta = 0$, then comlasso stops; 
\item[(C2)] sign condition: if $\bbeta(\lambda - \delta) \neq \bbeta(\lambda)$, then update $\cA$;
\item[(C3)] condition of piecewise constant in loss function: if
\bean
a(r_i(\lambda-\delta)) \neq a(r_i(\lambda)), \mbox{ or } b(r_i(\lambda-\delta)) \neq b(r_i(\lambda))  \mbox{ for some } I,
\eean
then update $H(\lambda - \delta)$;
\item[(C4)] dual feasibility condition: if $\mathcal{E}_{\lambda - \delta}(\bbeta(\lambda-\delta)) = \phi$, then update $\cA$ and $\cK$;
\item[(C5)] stationarity condition: if there exists $\bmu(\lambda - \delta)$, but
$$| \nabla_{j}  L(\bbeta(\lambda-\delta)) + (D\bmu(\lambda - \delta))_{j} | > (\lambda-\delta)$$ 
for some $j\in \cA^c$, then update $\cA$.
\eed

\noindent \rule[0.5ex]{\linewidth}{1pt}

In applying the updating rules, the signs of the active coefficients are determined by the inequality of each condition.
The determination of signs will be explained in the following subsection. If $\cA$, $\cK$, and $\mbox{sign}(\bbeta_{\cA})$ are obtained, then comlasso computes the direction of the solution path by solving \eqref{eq:linear} again and produces the solution path by increasing $\delta$ before one of the violations occurs. There is also the need to find the minimum $\delta>0$ that leads one of the violation events. We call the minimum $\delta$ the step size of the solution path.

\noindent \textbf{Remark}. In the classification problem,  \eqref{eq:KKT_active} in the KKT conditions is expressed as
\bean
\sum_{i=1}^n \left(2a(r_i(\lambda))\bx_{i\cA}\bx_{i\cA}^\top \bbeta(\lambda) \right) + b(r_i(\lambda))y_i\bx_{i\cA})  + (D\bmu(\lambda))_{\cA} + \lambda \mbox{sign}(\bbeta(\lambda)_{\cA}) &=& {\bf 0}_{|\cA|}.
\eean
Additionally, $H(\lambda)$ is same as that in the regression case. Therefore, equation \eqref{eq:linear} is also applied to find $({\bf b}(\lambda)^\top, {\bf m}(\lambda)^\top)^\top$ in the classification problem.

\subsection{Computation of the step size}

Let $\bbeta(\lambda)$ and $\bmu(\lambda)$ be the primal and dual solutions of \eqref{ob2}, respectively, and $(\bf{b}(\lambda)^\top , \bf{m}(\lambda)^\top)^\top$ the solution of \eqref{eq:linear}. It is easy to verify whether the first two conditions, (C1) and (C2), are still valid for $\delta \geq 0$. The violation of condition (C3) is explained by \cite{rosset2007piecewise} in detail, and we call this violation event ``hitting the knots.'' As such, we explain the dual feasibility (C4) and stationarity conditions (C5) to determine the step size and sign of the new activated coefficients.

First, we assume that step size is determined by the violation in the dual feasibility condition (C4). The dual feasibility condition is written as
    \bean
    d_{j}\mu_{k(j)}(\lambda -\delta) = -\nabla_{j} L(\bbeta(\lambda-\delta)) -\mbox{sign}(\beta_j(\lambda-\delta))(\lambda - \delta),
    \eean
    for $j\in \cA$, and
	\bean
	-(\lambda- \delta) - \nabla_j L(\bbeta(\lambda - \delta)) \leq d_{j} \mu_{k(j)}(\lambda - \delta) \leq 
	(\lambda - \delta) - \nabla_j L(\bbeta(\lambda - \delta)), 
	\eean
	for all $j \in \cA^c$. Here, $\nabla_j L(\bbeta(\lambda - \delta)) = \nabla_j L(\bbeta(\lambda)) + \sum_{i=1}^n \left(2a(r_i(\lambda))(\bx_{i})_j \bx_{i\cA}^\top {\bf m}(\lambda) \right)\delta, $, being a linear function of $\delta$.
	Therefore, the computation of step size is
	reduced to the problem of finding the maximum $\delta$ with
	$\cE_{\lambda-\delta}(\bbeta(\lambda-\delta)) \neq \phi$, 
	which can be solved by the following linear programming problem:
	\bea
	&&\max_{\delta \geq 0}~~ \delta \label{lp} \\  \nonumber
	&&\mbox{subject to } 
	\max_{j\in \cG_k}  \frac{\nabla_{j} L(\bbeta(\lambda-\delta))}{d_j} - 
	\frac{(\lambda - \delta) }{|d_j|}\\
	&&~~~~~~~~~~~~~~\leq
	\min_{j'\in \cG_k} \frac{\nabla_{j'} L(\bbeta(\lambda-\delta))}{d_{j'}} + 
	\frac{(\lambda - \delta) }{|d_{j'}|},
	\nonumber
	\eea
	for all $k \in \cK^c$. It is clear that the feasible set of problem \eqref{lp} is not empty. In fact, solving \eqref{lp} requires at most $\sum_{k=1}^K|\cG_k|(|\cG_k|-1)/2$ intersections of intervals.  
	
	Let $\delta^*$ be the solution of problem \eqref{lp}; then, there exist $k\in \cK^c$ and $(j,j')\in \cG_k$ so that
    \bea \label{eq:lm2}
    \frac{\nabla_{j} L(\bbeta(\lambda-\delta^*))}{d_j}-\frac{(\lambda - \delta^*) }{|d_{j}|} = \frac{\nabla_{j'} L(\bbeta(\lambda-\delta^*))}{d_{j'}}+\frac{ (\lambda - \delta^*) }{|d_{j'}|}, 
    \eea
    which is equal to $-\mu_{k}(\lambda - \delta^*)$. The violation of the dual feasibility condition means that $\bbeta(\lambda-\delta) + \delta {\bf m}(\lambda-\delta)$ for $\delta >\delta^*$ is not the solution of \eqref{eq:ob3} and
    \bean
    &&\nabla_j L(\bbeta(\lambda - \delta^*)) + d_j\mu_k(\lambda-\delta^*)  -  \mbox{sign}(d_j)(\lambda - \delta^*) = 0\\
    &&\nabla_{j'} L(\bbeta(\lambda - \delta^*)) + d_{j'}\mu_k(\lambda-\delta^*)  +  \mbox{sign}(d_{j'})(\lambda - \delta^*) = 0,    
    \eean
    for $k$ and $(j,j')$ in \eqref{eq:lm2}.
    The two equations above imply that $k$ and $(j,j')$ should be included in $\cK$ and $\cA$ to find a new direction of the solution path, respectively, and that the signs of the new activated coefficients are given by $\mbox{sign}(\beta_j) = - \mbox{sign}(d_j)$ and 
    $\mbox{sign}(\beta_{j'}) = \mbox{sign}(d_{j'})$.
  
   Next, we assume that step size is determined by the violation in the stationarity condition (C5) and let step size be $\delta^*$ again. In this case, $\bmu(\lambda - \delta^*)$ exists
   but, for some $j \in \cA^c$ and $k(j) \in \cK$,
   \bean
   \nabla_j L(\bbeta(\lambda - \delta^*)) + (D \bmu (\lambda - \delta^*))_j + (\lambda - \delta^*) = 0, 
   \eean
   or 
   \bean
   \nabla_j L(\bbeta(\lambda - \delta^*)) + (D \bmu (\lambda - \delta^*))_j - (\lambda - \delta^*) = 0.
   \eean
   To find a new direction, we should include index $j$ in $\cA$ and let $\mbox{sign}(\beta_j(\lambda - \delta^*)) = 1$
   for the former case and $\mbox{sign}(\beta_j(\lambda - \delta^*)) = -1$ for the latter.

\subsection{Uniqueness of the solution path}
    As previously mentioned, the comlasso algorithm consists of the initialization, determination of the signed active set, and computation of step size. We summarize comlasso as follows:

\noindent \rule[0.5ex]{\linewidth}{1pt}
\bc
{\bf comlasso algorithm}
\ec

\bed
\item Initialization 
    \bed
    \item Compute $\lambda_{\max}$ from \eqref{lambda_max} and let $\lambda = \lambda_{\max}$ and $\bbeta(\lambda) = {\bf 0}_p$;
    \item Initialize $\cA$ and $\cK$ and compute the corresponding dual parameter $\mu_k$ for $k \in \cK$;
    \item Assign the sign of the active coefficients $\beta_j(\lambda)$ for $j \in \cA$.
    \eed
\item Repeat
	\bed
	\item Compute $\bf{b}(\lambda)$ and ${\bf m}(\lambda)$ by solving \eqref{eq:linear};
	\item Compute $\delta^* = \min(\delta_1, \delta_2, \delta_3, \delta_4, \delta_5)$ based on the updating rules in Section 2.2:
	\ben
	\item[(a)] $\delta_1 = \lambda$;
	\item[(b)] $\delta_2 = \sup\{\delta \geq 0: \mbox{sign}(\beta_j(\lambda - \delta)) = \mbox{sign}(\beta_j(\lambda)) \mbox { for all } j \in \cA\}$;
	\item[(c)] $\delta_3$ is the smallest $\delta$ of ``hitting the knot;''
	\item[(d)] $\delta_4$ is computed by \eqref{lp};
	\item[(e)] $\delta_5 = \{\delta \geq 0: |\nabla_j L(\bbeta(\lambda - \delta)) + (D \bmu (\lambda - \delta))_j| \leq (\lambda - \delta)  \mbox{ for all }j \in \cA^c, k(j)\in \cK \}.$
	\een
	\item Let $\bbeta_{\cA}(\lambda - \delta^*) = \bbeta_{\cA}(\lambda) + \delta^* {\bf b}(\lambda)$ and $\bmu_{\cK}(\lambda - \delta^*) = \bmu_{\cK}(\lambda) + \delta^* {\bf m}(\lambda)$ and update $\lambda \leftarrow \lambda - \delta^*$,
    $\cA$, $\cK$, and the sign of $\bbeta_{\cA}(\lambda)$;
    \item If $\delta^* = 0$, then stop. 
	\eed

\eed
\noindent \rule[0.5ex]{\linewidth}{1pt}
 
    As a special case of \eqref{ob2}, 
    we consider the problem \eqref{eq:comlasso}.
   Let $j$ and $j'\in \{1, \cdots,p\}$ be the index satisfying
$X_j^\top \by \leq X_l^\top \by$ for all $l\neq j$ and $X_{j'}^\top \by \geq X_l^\top \by$ for all $l\neq j'$. Then, the initializing regularization parameter is $\lambda_{\max} = (X_{j'} - X_{j})^\top \by/2$ and 
 the dual parameter associated with the equality constraint is uniquely given by $\mu = (X_j+X_{j'})^\top\by/2$. The following theorem shows that the solution path in problem \eqref{eq:comlasso} is unique and, thus, comlasso pursues the unique solution path. 
{\theorem If any $n\times n$, the submatrix of $X$ has full rank; for $\lambda >0$, there is the unique solution of the problem \eqref{eq:comlasso} and comlasso provides the solution path for $\lambda >0$. 
}  

\begin{proof}
{
 We fix $\lambda>0$ and denote the dual solution of \eqref{eq:comlasso} by $\hat \mu \in \mathbb{R}.$ First, we show that $\hat \mu$ is unique. Let $\hat \bbeta$ and $\tilde \bbeta$ be the solution of \eqref{eq:comlasso}. Additionally, we set $\tilde \mu$ as another dual solution and let 
$\cA_{\hat \mu} = \{j: |X_j^\top (y - X\hat \bbeta) + \hat \mu| = \lambda \}$ and $\cA_{\tilde \mu} = \{l: |X_l^\top (y - X\tilde \bbeta) + \tilde \mu|= \lambda \}$.
From the strict convexity of the objective function, $X\hat \bbeta = X \tilde \bbeta$ and $X^\top (y - X\hat \bbeta) =  X^\top (y -  X\tilde \bbeta)$ (see Lemma 1 in \cite{tibshirani2013lasso}). The dual function is always concave, so that $t \tilde \mu + (1-t) \hat \mu$ for all $t \in [0,1]$ also represent the dual solution. Assuming that $\cA_{\hat \mu} \cap \cA_{\tilde \mu}$ is not empty, 
\bean
t \hat \mu + (1-t) \tilde \mu = X_j^\top(Y- X\hat \bbeta) - \lambda \mbox{sign}(\hat \beta_j),
\eean
for $j \in \cA_{\hat \mu} \cap \cA_{\tilde \mu},$ which is constant. Therefore, $\cA_{\hat \mu} \cap \cA_{\tilde \mu}$ must be empty if $\hat \mu \neq \tilde \mu$. However, we know that $t\hat \bbeta + (1-t) \tilde \bbeta$ for $t\in (0,1)$ is also the primal solution of \eqref{eq:comlasso}, which is a contradiction that the intersection of the active sets of primal solutions $\hat \bbeta$ and $t\hat \bbeta + (1-t) \tilde \bbeta$ must be empty. That is, $\hat \mu$ is unique if $\cA_{\hat \mu}$ is not empty.

From the uniqueness of the dual solution, the corresponding $\hat \bbeta$ is unique because any $n\times n$ submatrix of $X$ has full rank (see Theorem 5 in \cite{osborne2000lasso}). This result implies that comlasso pursues the unique solution of \eqref{eq:comlasso} for $\lambda >0$. 
}
\end{proof}

\medskip

\section{Numerical examples}
\subsection{Toy example analysis}
The comlasso algorithm can be applied to the compositional regression model estimation with adaptive lasso penalty \citep{zou2006} through simple reparametrization. We applied the adaptive lasso to sand data analysis \citep{Aitchison:1986:SAC:17272} as a toy example of compositional regression. 
The sand data consist of 39 sediment samples, collected at different water depths. The sediment components are sand, silt, and clay, measured in percentages for each sample. Water depth is the response variable and the others are predictors in the compositional regression. The weights for adaptive penalty are the inverses of non-regularized estimates. We compare the solution paths and the regression coefficients of the lasso and adaptive lasso estimates chosen by Bayesian information criteria (BIC). We set the degree of freedom in BIC as the number of active coefficients subtracted from 1.

\begin{figure}[!hbpt]
\begin{center}
\includegraphics[width = 0.75 \textwidth, height=8cm]{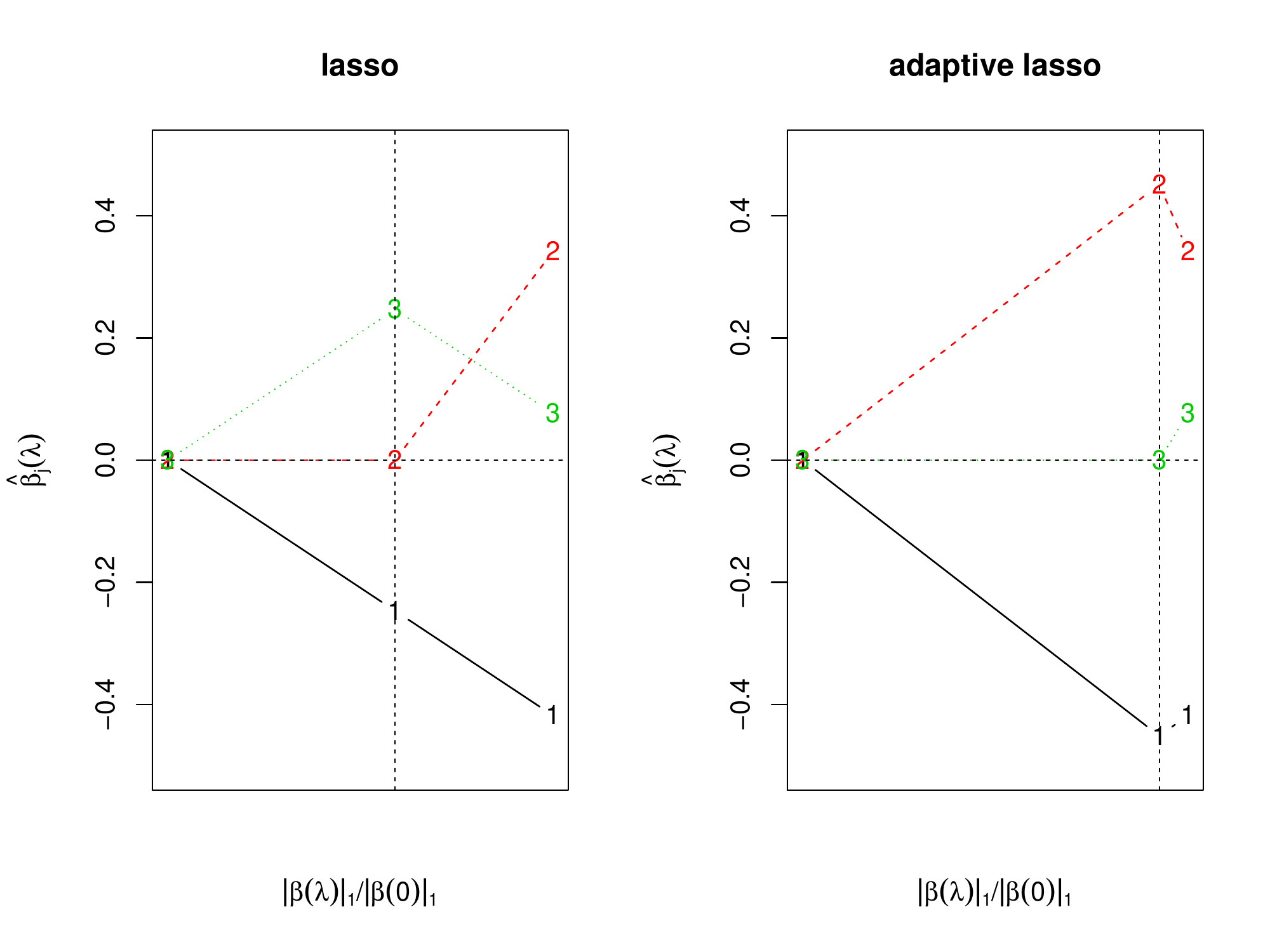}
\caption{Solution paths of lasso (left) and adaptive lasso (right). The numbers on the solution paths indicate sand, silt, and clay in this order. \label{fig:sandpath}}
\end{center}
\end{figure}

Figure \ref{fig:sandpath} shows the solution paths of lasso (left) and adaptive lasso (right). The vertical line denotes the selected model using BIC. The non-regularized estimates corresponding to the predictor, clay, are relatively small. A large penalization on the predictor under adaptive lasso leads different solution paths from that of lasso.

\subsection{Speed comparison}

Comlasso adopts the idea of LARs \cite{efron2004least} and the piecewise linear toolbox \citep{rosset2007piecewise} that pursue the solution path under lasso. Compared with LARs, the computational cost of the comlasso algorithm is not significantly more expensive. Only $|\cK|$ more Lagrangian variables are used in the linear system to find the solution path directions and
a maximum of $\sum_{k\notin \cK} p_k(p_k-1)/2$ additional arithmetic computations are required to obtain the step size. Compared with genlasso \citep{tibshirani2011solution}, comlasso has a computational advantage in a high-dimensional regression model. In fact, genlasso was originally developed to solve the regression problem with the full ranked design matrix, so that it employs an additional $l_2$ penalty when $p>n$. The $l_2$ penalty function is incorporated into the design matrix for the full rank design, which leads to a drastic increase in computational cost, especially when $p$ is significantly larger than $n$ (see Section 7 in \cite{tibshirani2011solution}). However, comlasso does not require the full ranked design matrix, nor suffer from such a computation problem.

We consider an optimization problem with quadratic loss and perform two simulations to compare the computational speeds of genlasso and comlasso. The simulation setting is similar to \cite{lin2014variable}. Here, $\bx_i \in \mathbb{R}^p$ for $i = 1, \cdots, n$, are generated from the logarithm of the logistic normal distribution with $(\bfeta, \Sigma)$, where $\bfeta_{\cG_k} = \log(p_k/2) {\bf 1}_{p_k}$ for $k = 1, \cdots, 5$,   $\bfeta_{\cG_k} = {\bf 0}_{p_k}$ for $6 \leq k \leq K$, and $\Sigma$ is the block diagonal covariance matrix, whose diagonal block is given by $\Sigma_{k}$ for $k = 1, \cdots, K$ with $(\Sigma_k)_{jl} = 0.5^{|j-l|}$ for $1\leq j,l\leq p_k$. The regression parameters are given by $\bbeta_{\cG_1} = (1, -0.8,0.6,0,0,-1.5,-0.5,1.2, 0 \cdots, 0)^\top\in \mathbb{R}^{p_1}$ and $\bbeta_{\cG_k}  ={\bf 0}_{p_k}$ for $k = 2, \cdots, K$, and $\epsilon_i$s are independently generated from $N(0,0.5^2).$

For a fair comparison of computational time, we control for the maximum number of kinks in the solution path. Due to the use of the $l_2$ penalty function in the high-dimensional cases, genlasso produces less sparse solutions with more kinks, even when the degree of freedom achieves maximum in the primal problem. To prevent redundant computations for genlasso, we bind the maximum number of kinks for which comlasso achieves the maximum degree of freedom in the primal problem. Additionally, we do not include the computation time of reparametrization required for genlasso.
We use R for the comlasso algorithm (see  \url{https://github.com/jenjong/ComLasso}) and implement genlasso and comlasso using a PC with 3.8 Ghz CPU and 16 GB memory under a Win 10 operating system. Computation time is measured by 20 repetitions.

In the first simulation, we let $K = 1$ and vary $p$ and $n$. Table \ref{tab:table1} presents the average computation times and standard errors for each $p$ and $n$. Comlasso is efficient compared to genlasso, especially when $p$ is large. In addition, comlasso is less affected by $p$ than genlasso. This may be because the main computational burden in comlasso arises in the matrix inversion to determine direction. 

\begin{table}[h]
    \centering
        \caption{Average computation time and standard errors (between parentheses)}
    \label{tab:table1}
    \begin{tabular}{cc c c|c c c}  \hline
        &\multicolumn{3}{c|}{$n=50$} & \multicolumn{3}{c}{$n=100$}\\
    $p$ & 200 & 500 & 1000 & 200 & 500 & 1000\\ \hline 
        comlasso &  0.23 (0.01) & 0.22 (0.00) & 0.22 (0.00) & 0.52 (0.01) & 0.54 (0.01) & 0.55 (0.01)\\
        genlasso & 0.27 (0.01) & 3.25 (0.06) & 19.23 (0.33) & 0.43 (0.01) & 4.18 (0.07) & 30.44 (0.34)  \\ \hline
    \end{tabular}

\end{table}

In the second simulation, we fix $p = 1000$ and vary $n$ and $K$. We let $p_k=p/K$ for $k = 1, \cdots, K$. Table \ref{tab:table2} summarizes the simulation results. Comlasso still performs well in the high-dimensional case. As $K$ increases, more computations are required for checking the violation of the KKT conditions \eqref{eq:KKT_inactive}, so that comlasso becomes slightly slower to produce the solution path. Genlasso seems to be faster as $K$ increases, that is, the dimension of $\mbox{null}(D^\top)$ increases for a fixed $n$.

\begin{table}[h]
    \centering
        \caption{Average computation time and standard errors when $p =1000$}
    \label{tab:table2}
    \begin{tabular}{cc c c|c c c}  \hline
        &\multicolumn{3}{c|}{$n=50$} & \multicolumn{3}{c}{$n=100$} \\ 
        $K$ & 20 & 50 & 100 & 20 & 50 & 100\\ \hline 
        comlasso & 0.26 (0.01) & 0.23 (0.00) & 0.24 (0.01) & 0.67 (0.02) & 0.83 (0.02) & 0.90 (0.02)\\
        genlasso & 21.93 (0.48) & 20.57 (0.36) & 18.84 (0.36) & 35.16 (0.53) & 36.75 (0.67) & 32.49 (0.55) \\ \hline
    \end{tabular}

\end{table}

\textbf{Remark}. (comparison of sparsity) Since genlasso \citep{tibshirani2011solution} uses the $l_2$ penalty function, its solution is less sparse than the optimal solution. The sparse solution can be obtained by proper thresholding and most calibrated solutions are close to those of comlasso, at least in our experiments. However, there are no studies for the determination of the threshold level in this case.

\subsection{Real data analysis}

We apply the comlasso algorithm to estimate a
classification model that predicts the incidence of bloodstream infections (BSI)
in patients. We use the microbiome dataset \citep{Montassier2016}, which
consists of 11 BSI patients and 17 non-BSI patients and amounts to 3837 operational taxonomic units (OTUs)
of microbiomes, that is, pragmatic proxies for microbial ``species,'' identified by DNA sequencing.
The 3837 OTUs belong to one of 111 genera, which in turn belong to 6 phyla. 
In the analysis, we exclude the two phyla and the corresponding genera and OTUs 
because they only have a single OTU.
As a result, the 3833 OTUs belonging to the 107 genera and 4 phyla are used for further analysis.

We aggregate the amounts of OTUs in each genus and use them as covariates. That is,
the data consist of one binary response and 107 genus-level amounts of OTUs.
From this dataset, we compose two compositional datasets. The first one is to take the proportions
of the 107 genus-level amounts of the OTUs (i.e. $K=1$ and $p=107$). 

The second dataset considers the proportions between genera belonging to the same phylum. 
There are four types of phyla (Actinobacteria, Bacteroidetes, Firmicutes, and Proteobacteria), 
Which include 20, 10, 62, and 15 genera, respectively, and let the index set of genera denoting their phylum levels be $\cG_k\subset \{1, \cdots, p\}$ for $k= 1, \cdots, 4$. Therefore, the dimensions of the compositional predictors are
$K=4$, and $(p_1,p_2,p_3,p_4) = (20,10,62,15)$.
The averages of genus-level aggregated OTUs in each phylum are $45.14$, $118.71$, $50.11$, and $3.28,$, respectively,
being quite different from each other. Hence, the proportions separately calculated for each phylum in the
second dataset are quite different from the proportions among all genera in the first dataset.
Note that the compositional regression model for the second dataset is a special case
of the compositional regression model for the first dataset with additional constraints on the regression
coefficients of each phylum. Therefore, using the proportions in each phylum as compositional predictors can reflect the information about the phylogenetic tree into the classification model.

We use the quadratic hinge loss (see the Appendix) to measure the predictive performance of the estimated models by 
leave-one-out cross validation. Additionally, we apply the stability selection procedure \citep{meinshausen2010stability} to assess the importance of predictors. 

\begin{figure}
    \centering
    \includegraphics[width = 0.32 \textwidth, height=4.6cm]{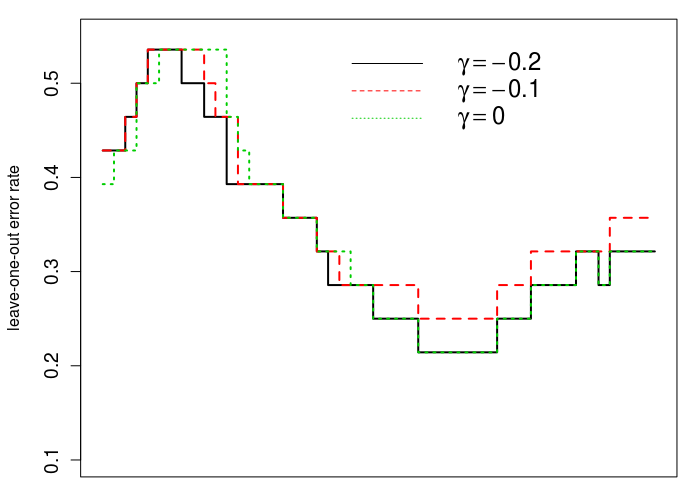}
   \includegraphics[width = 0.32 \textwidth, height=4.6cm]{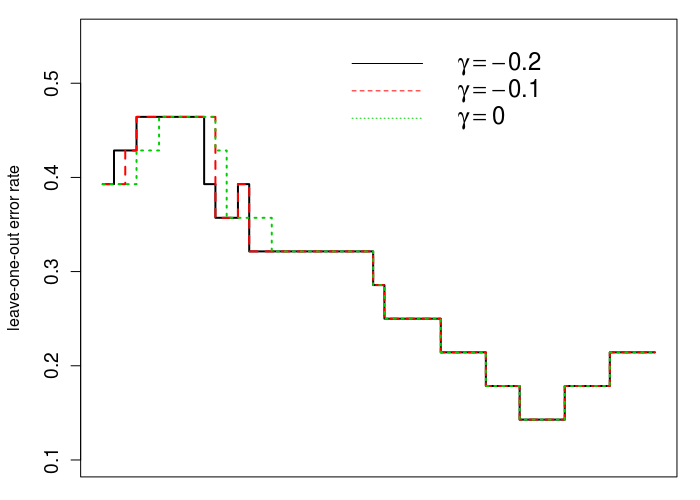} 
    \includegraphics[width = 0.32 \textwidth, height=4.6cm]{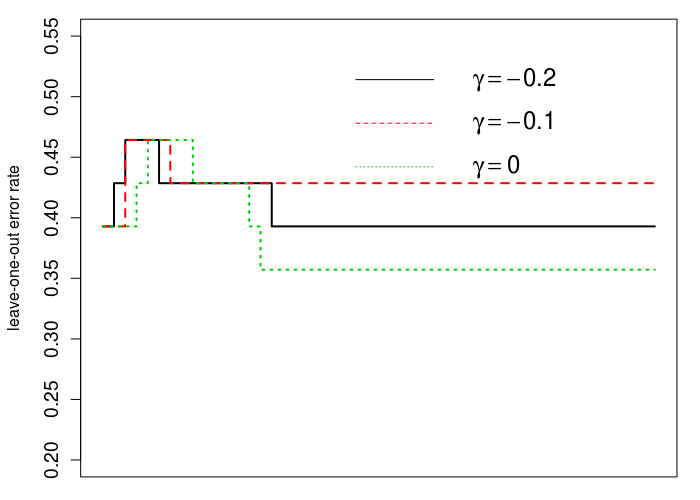}   
    \caption{Leave-one-out errors for comlassoA (left), comlassoB (middle), and non-reguralized model (right): $\gamma$ is the parameter in the squared hinge loss}
    \label{fig:CV}
\end{figure}

Figure \ref{fig:CV} presents the leave-one-out errors of the three models, comlassoA (for the first dataset), comlassoB (for the second dataset,) and the non-regularized model \citep{Montassier2016}, where the model is fitted with the selected predictors by the Wilcoxon test in the first dataset. The figure illustrates that comlassoA and comlassoB show lower cross validation errors than the non-regularized model and 
that comlassoB tends to outperform comlassoA. Additionally, the choice of $\gamma$ in the 
quadratic hinge loss does not affect the results significantly. 

\begin{figure}[!htbp]
\centering
\includegraphics[width=0.49 \textwidth]{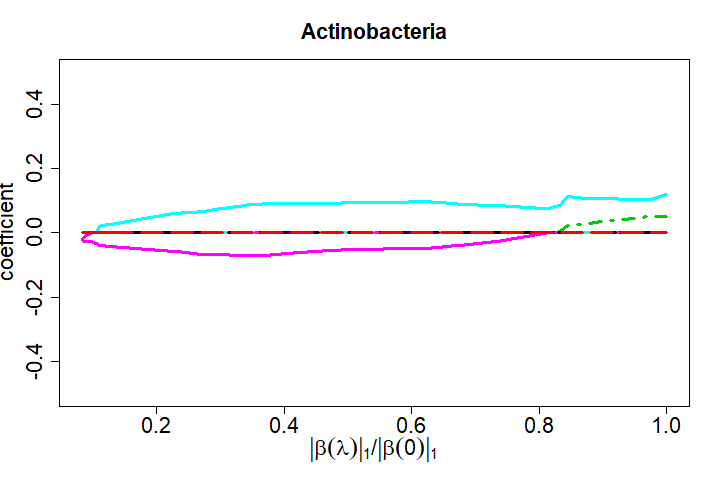}
\includegraphics[width=0.49 \textwidth]{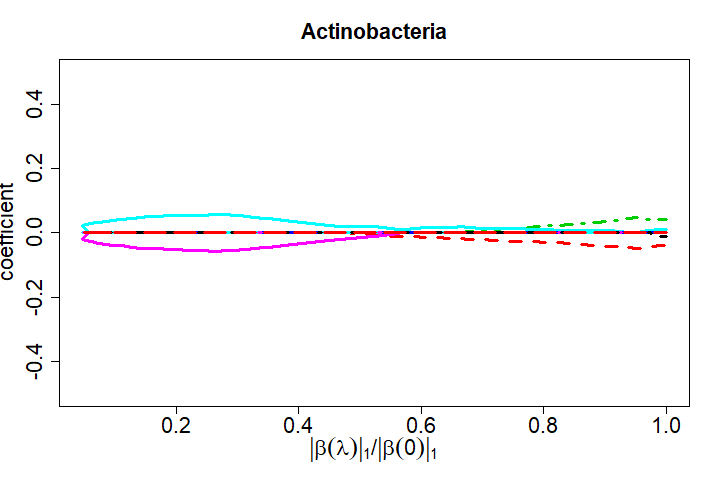} \\
\includegraphics[width=0.49 \textwidth]{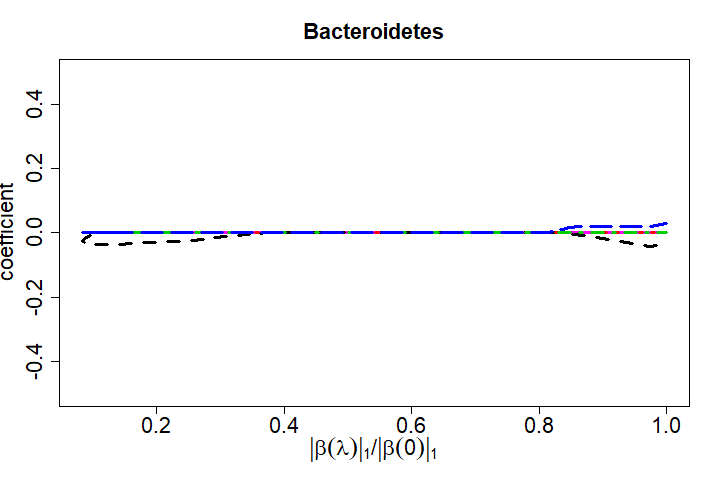}
\includegraphics[width=0.49 \textwidth]{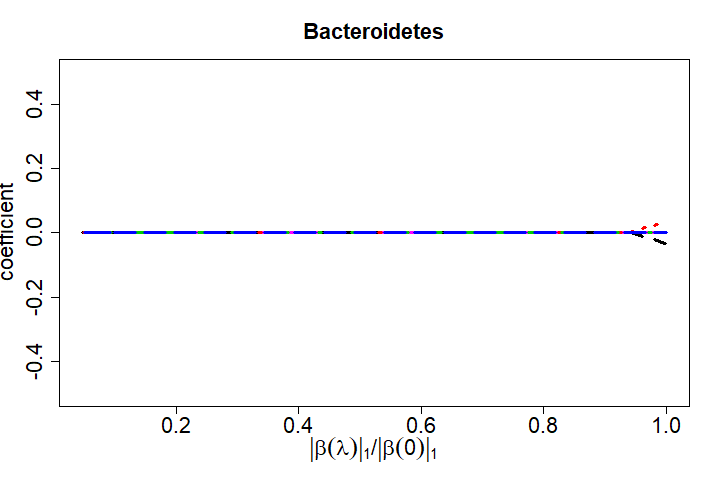} \\
\includegraphics[width=0.49 \textwidth]{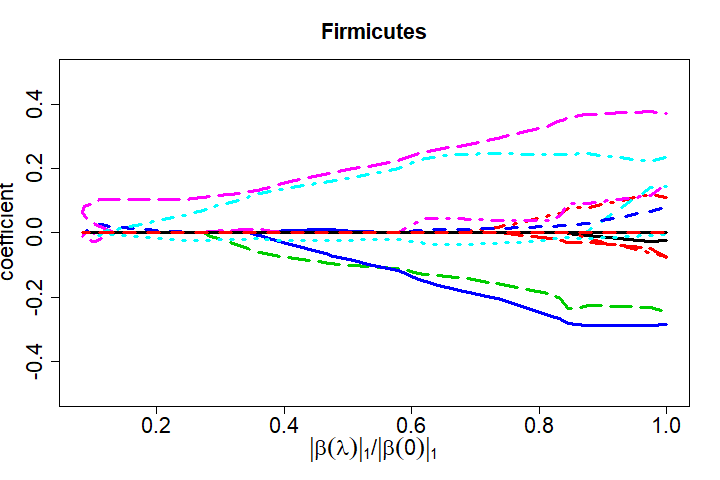}
\includegraphics[width=0.49 \textwidth]{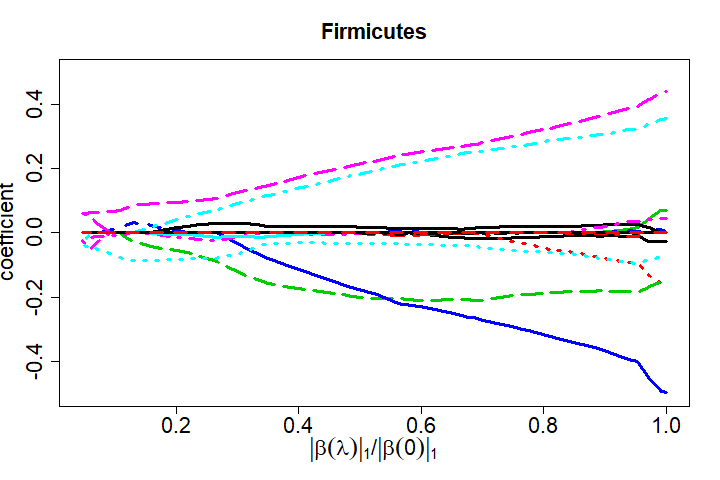} \\
\includegraphics[width=0.49 \textwidth]{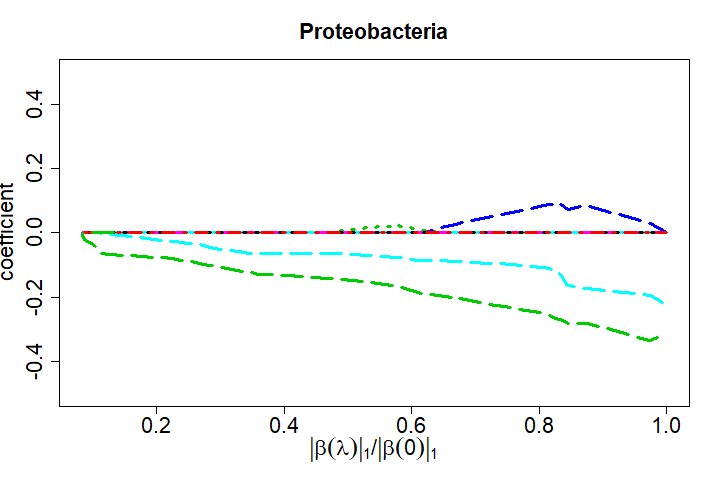}
\includegraphics[width=0.49 \textwidth]{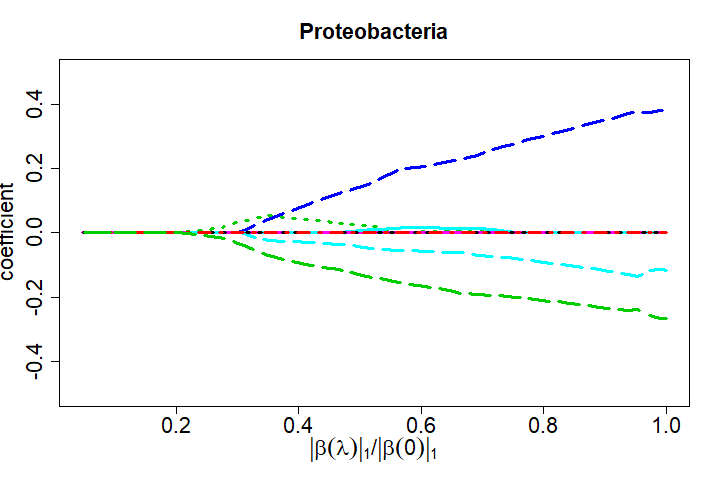} 
\caption{Solution paths of comlassoA (left-hand side graphs) and comlassoB (right-hand side graphs) for $\gamma=-0.2.$ } \label{fig:bsi_solpath}
\end{figure}

Figure \ref{fig:bsi_solpath} draws the entire solution paths for each phylum of comlassoA and 
comlassoB for $\gamma=-0.2$. The horizontal axis denotes the value of $\|\bbeta_{\cG_k}(\lambda)\|_1/\|\bbeta_{\cG_k}(\lambda_{\max})\|_1$.
 Figure \ref{fig:bsi_solpath} also shows substantial differences among the estimated coefficients of the genera in the Proteobacteria phylum group. Specifically, for a large $\lambda$, that is, for a small $\|\bbeta(\lambda)\|_1/\|\bbeta(\lambda_{max})\|_1$, comlassoB gives more sparse coefficients at the phylum level than comlassoA. As previously pointed out, the average number of OTUs in each genus in the Proteobacteria group is significantly lower than the others. This means that the relative ratios seriously depend on the set where ratios are computed. Therefore, the coefficients corresponding to the Proteobacteria group are activated in an earlier step in comlassoA than comlassoB. This can be explained by a normalization of composite predictors, which reflects the structure of the phylum as prior knowledge of features.


Finally, to evaluate the importance of predictors we use the stability selection 
procedure \citep{meinshausen2010stability}. Specifically, we use 1000 subsamples for half of the samples and fit the randomized lasso estimates with weakness parameter $0.5$ for each subsample. The regularization parameter of the randomized lasso is chosen by five-fold cross validation. Comlasso is applied to implement the randomized lasso algorithm.
Table 1 summarizes the 10 genera with the highest selection probabilities
in comlassoA and comlassoB, as well as the top-10 genera with the smallest p-values in the Wilcoxon test. 
Table 1 also shows that Lactobacillus was selected as the most important genus for both comlassoA and comlassoB. 
The Wilcoxon test also indicates that Lactobacillus is significant in the top-10 rank in terms of $p$-value. 
Multiple literatures showed Lactobacillus can cause serious infections of the bloodstream \citep{Cannon2005,Salminen2002}.

Table \ref{tbl:bsi} indicates that the multiple genus in the Proteobacteria phylum are selected by comlassoA and Wilcoxon test. However, comlassoB does not select any genus included in the phylum within the top-10 ranked predictors. As previously discussed, this selection pattern of comlassoB is partly owed to a normalization in compositional data. Considering predictive performance, we may conclude that it is worth investigating additional biological evidence
based on the results of comlassoB under larger samples.

\begin{table}[!htbp]
\small
\centering \caption{Genera lists using three methods: Selection
probability for comlassoA and comlassoB and p-value for the marginal
Wilcoxon test.
Between parentheses, the letters indicate the phylum level as follows: 
A stands for ``Actinobacteria;'' B for ``Bacteroidetes;'' F for ``Firmicutes;'' and P for ``Proteobacteria.'' 
} \label{tbl:bsi}
\begin{tabular}{cllllllll} \\ \hline
& {comlassoA } & {comlassoB} & {Wilcoxon test} \\
No. & Genera (sel. Prob.)         & Genera (sel. Prob.)               & Genera (p-value) \\ \hline
  1 & F. Lactobacillus (0.744)   & F. Lactobacillus (0.645)         & F. Faecalibacterium (0.001) \\ 
  2 & A. Bifidobacterium (0.468) & F. Phascolarctobacterium (0.357) & P. Sutterella (0.003) \\ 
  3 & P. Sutterella (0.418)      & F. Faecalibacterium (0.334)      & F. Oscillospira (0.004) \\ 
  4 & B. Parabacteroides (0.402) & F. Christensenella (0.324)       & F. Dehalobacterium (0.006) \\ 
  5 & P. Desulfovibrio (0.360)   & F. Roseburia (0.314)             & P. Oxalobacter (0.006) \\ 
  6 & F. Enterococcus (0.341)    & A. Eggerthella (0.297)           & P. Desulfovibrio (0.008) \\ 
  7 & F. Turicibacter (0.328)    & A. Bifidobacterium (0.274)       & B. Butyricimonas (0.022) \\ 
  8 & A. Eggerthella (0.290)     & B. Prevotella (0.260)            & F. Christensenella (0.022) \\ 
  9 & A. Collinsella (0.270)     & F. Clostridium (0.250)           & B. Parabacteroides (0.053) \\ 
 10 & F. Roseburia (0.264)       & F. Enterococcus (0.240)          & F. Lactobacillus (0.053) \\ \hline
\end{tabular}
\end{table}

\section{Concluding remarks}

We developed a solution path algorithm for an $l_1$ regularized regression model with a compositional predictor in the high-dimensional case.
Comlasso generalizes the equiangular direction of the LARs on a specific linear space and directly produces the entire solution path under the primal problem. Compared with genlasso, comlasso has the advantage of not only numerical precision but also computational efficiency in the high-dimensional regression model. 
Moreover, comlasso is easy to be extended to the regularized regression problem
with almost quadratic loss, such as the Huberized loss and expectile loss functions. Because comlasso does not require continuously twice differentiability of the loss function, 
it can provide a new, efficient method for optimizing \eqref{ob2}, where the solution path algorithm of \cite{gaines2018algorithms} is not covered.

Additionally, we elaborated a way of monitoring the feasibility of the dual parameters corresponding to the equality constraints. In our case, verifying the condition requires only finite intersections of half intervals. Although we assume a special structure of the considered equality constraint, the proposed method to monitor dual feasibility can be extended to a general linear constraint, in which step size is computed by linear programming. 

As an application of comlasso, we used the $l_2$ hinge loss for the classification problem and analyzed the microbiome dataset. We also considered two different types of constraints and compared the predictive performances of the considered models. The numerical results indicate that the regularized models show better performance than the non-regularized one. Additionally, we implemented stability selection to evaluate the importance of predictors.



\appendix
\section{Appendix}

\subsection{Examples}

{\Example \label{EX:loss1}{ Loss function in regression } 
\bed
\item Quadratic loss is given by $l(y,\bx^\top \bbeta)=r^2/2,$ and, in this case, $a(r)=1/2$ and $b(r) = c(r)=0$.

\item The asymmetric $l_2$ loss with $h \in (0,1)$ is given by
\bean
l(y,\bx^\top \bbeta) = \begin{cases}
(1-h)r^2/2 & \mbox{ if } r \le 0, \\
h r^2/2 & \mbox{ otherwise},
\end{cases}
\eean
Where the corresponding constants are piecewise defined as
$a(r)=(1-h), b(r)=c(r)=0$ for $r \leq 0,$
and $a(r)= h, b(r)=c(r)=0$ for $r> 0 $ \citep{aigner1976estimation}.

\item The Huberized loss function with a fixed knot $h$ is given by
\bean
l(y,\bx^\top \bbeta)= \begin{cases}
r^2/2 & \mbox{ if } |r| \le h, \\
h|r|-h^2/2 & \mbox{ otherwise},
\end{cases}
\eean
where the corresponding constants are piecewise defined as
$a(r)=1/2, b(r)=c(r)=0$ for $|r|\leq h,$
and $a(r)=0, b(r)=h {\rm sign}(r), c(r)=-h^2/2$ for $|r|>h $ \citep{rosset2007piecewise}.

\eed

}
  
{\Example { Loss function in classification} \label{loss:hinge}

\bed
\item The squared hinge loss function is given by $l(y,\bx^\top \bbeta)=\big(\max(0,1-r)\big)^2/2$, in which 
$a(r)=b(r)=c(r)=0$ for $r>1$ and $a(r)=1/2, b(r)=-1,c(r)=1/2$ for $r<1$ \citep{lee2013study}.
\item The Huberized loss function with a fixed knot $h (\le 1)$ is given by
\bean
l(y,\bx^\top \bbeta)= \begin{cases}
\big(\max(0,1-r)\big)^2/2             & \mbox{ if } r \ge h, \\
-(1-h)r+(1-h^2)/2 & \mbox{ otherwise},
\end{cases}
\eean
in which $a(r)=0$, $b(r)=0$, $c(r) = 0$ for $r \ge 1,$ and $a(r)=1/2$,  $b(r)=-1$, $c(r) = 0$ for $h \le r \le 1,$ $a(r)=0$, $b(r)=h-1$, $c(r) = 0$ for $r \le h$ \citep{rosset2007piecewise}.
\eed
}
\bibliographystyle{apalike}
\bibliography{comlasso_rev}

\end{document}